\documentclass[letterpaper, 10 pt, journal, twoside]{ieeetran}  

\IEEEoverridecommandlockouts                              

\usepackage{graphics} 
\usepackage{subfiles}
\usepackage{csquotes}
\usepackage{gensymb}
\usepackage{pifont}
\usepackage{textcomp}
\usepackage[table]{xcolor}
\usepackage{todonotes}
\usepackage{pdfpages}
\usepackage{blindtext}
\usepackage{soul}
\usepackage[breaklinks,colorlinks]{hyperref}
\usepackage{graphicx}
\usepackage{mathtools}
\usepackage{multirow}
\usepackage{booktabs}
\usepackage[hang,flushmargin]{footmisc}

\graphicspath{{images/}{../images/}}
\usepackage[font={footnotesize}]{caption}
\usepackage[export]{adjustbox}
\usepackage[resetlabels]{multibib}
\newcites{New}{References}
\usepackage{cuted}

\usepackage{amsmath} 
\usepackage{amssymb} 
\usepackage{amsfonts}
\usepackage{bm} 
\usepackage{siunitx}
\usepackage{cleveref}
\usepackage{cancel}
\usepackage[inline]{enumitem}

\usepackage{subcaption}
\usepackage{xcolor}

\usepackage{microtype}

\renewcommand{\eqref}[1]{Eq.~(\ref{#1})}

\title{\LARGE \bf
Hyperspectral Adapter for Semantic Segmentation with\\Vision Foundation Models
}

\author{
Juana Valeria Hurtado, Rohit Mohan, and Abhinav Valada 
\thanks{Department of Computer Science, University of Freiburg, Germany.}%
\thanks{This work was financed by the Baden-Württemberg Stiftung gGmbH within the programm “Automone Robotik”. Rohit Mohan was supported by the Bosch Research collaboration on AI-driven automated driving.}
}

\begin{document}

\maketitle
\pagestyle{empty}
\thispagestyle{empty}

\begin{abstract}
Hyperspectral imaging (HSI) captures spatial information along with dense spectral measurements across numerous narrow wavelength bands. This rich spectral content has the potential to facilitate robust robotic perception, particularly in environments with complex material compositions, varying illumination, or other visually challenging conditions. However, current HSI semantic segmentation methods underperform due to their reliance on architectures and learning frameworks optimized for RGB inputs. In this work, we propose a novel hyperspectral adapter that leverages pretrained vision foundation models to effectively learn from hyperspectral data. Our architecture incorporates a spectral transformer and a spectrum-aware spatial prior module to extract rich spatial-spectral features. Additionally, we introduce a modality-aware interaction block that facilitates effective integration of hyperspectral representations and frozen vision Transformer features through dedicated extraction and injection mechanisms.
Extensive evaluations on three benchmark autonomous driving datasets demonstrate that our architecture achieves state-of-the-art semantic segmentation performance while directly using HSI inputs, outperforming both vision-based and hyperspectral segmentation methods.  We make the code available at \url{https://hsi-adapter.cs.uni-freiburg.de}.
\end{abstract}

\section{Introduction}
\label{sec:introduction}

Advancing robot perception in complex environments requires an understanding of detailed scene characteristics beyond surface appearance~\cite{mohan2024progressive}. Hyperspectral imaging (HSI), which captures reflectance information across tens to hundreds of narrow spectral bands, provides material-specific information that is significantly richer than conventional RGB imaging~\cite{guerri2024deep}. By encoding distinctive spectral signatures, HSI enables fine-grained discrimination of objects and materials, offering the potential to enhance perception tasks~\cite{mohan2024panoptic, luz2024amodal} under challenging conditions such as varying illumination, occlusions~\cite{sekkat2024amodalsynthdrive}, scene clutter~\cite{mohan2023syn}, and complex material compositions. These properties make HSI particularly promising for autonomous driving, where reliable perception in diverse and visually complex environments is essential.

\begin{figure}
    \centering
    \includegraphics[width=0.9\linewidth]{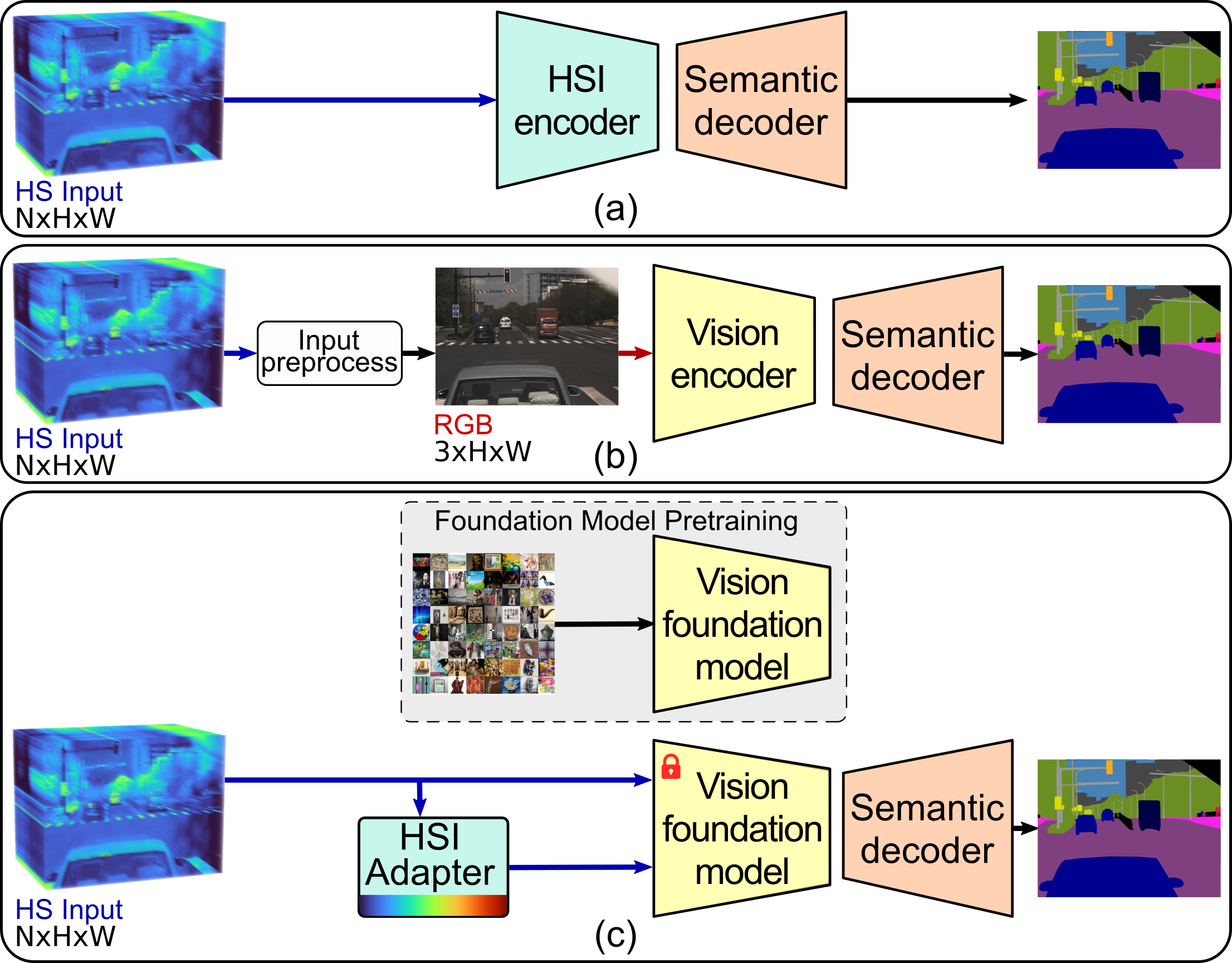}
    \caption{Different approaches for hyperspectral semantic segmentation. (a)~Traditional approaches train HSI encoders with high parameter counts from scratch. Due to the lack of large-scale annotated hyperspectral datasets, this results in poor generalization.
(b)~Preprocessing the hyperspectral input allows using vision encoders with pseudo RGB images, but ignores the rich spectral information present in hyperspectral data.
(c)~Our paradigm adapts a pretrained frozen vision foundation model to the hyperspectral domain via a lightweight HSI adapter, effectively bridging spectral and vision representations.}
    \label{fig:coverimages}
    \vspace{-0.3cm}
\end{figure}

Hyperspectral semantic segmentation aims to classify each pixel of an image into predefined categories based on HSI inputs. Despite its potential, semantic segmentation with hyperspectral data remains a challenging problem. Most segmentation models are tailored for RGB images and fail to fully leverage the spectral richness of HSI. Standard vision semantic segmentation architectures, although proven powerful for extracting spatial semantics from RGB data, are not inherently equipped to model the complex inter-channel dependencies of hyperspectral inputs. As a result, models pretrained on RGB datasets often underperform when directly applied to hyperspectral inputs~\cite{theisen2024hs3,valada2016convoluted,mohan2023neural}. Although pseudo-RGB approaches can partially adapt existing models, they overlook the spectral diversity inherent in HSI and fail to address critical challenges such as high dimensionality, limited labeled data, and increased computational costs~\cite{guerri2024deep}. We illustrate the difference between these approaches in Fig.~\ref{fig:coverimages}.

Vision foundation models trained on large-scale datasets generalize well across domains~\cite{kappeler2024few}. However, their full finetuning is computationally prohibitive, and linear probing lacks capacity for complex tasks such as hyperspectral segmentation. Adapter tuning provides an efficient alternative by inserting lightweight modules into a frozen backbone, allowing for task-specific adaptation without retraining the entire model. However, existing adapters are designed for RGB inputs and, when applied to HSI, they face initialization sensitivity and fail to capture spectral dependencies, limiting their effectiveness.

To address these limitations, we propose the HSI-Adapter, a modular architecture that enables leveraging strong pretrained vision foundation models with HSI inputs for hyperspectral semantic segmentation. Our approach introduces a spectral transformer and a spectral-enhanced spatial prior module to jointly extract rich spectral and spatial features from hyperspectral inputs. To bridge the modality gap between hyperspectral and RGB representations, we propose a modality-aware interaction block that enables effective bidirectional feature exchange between hyperspectral features and foundational vision features. This topology preserves the strengths of the pretrained foundation model while incorporating the unique characteristics of hyperspectral data. Extensive experiments on HSI-DriveV2~\cite{gutierrez2023hsi}, HyperspectralCityV2.0 (HCV2)~\cite{winkens2017hyko}, and HyKo2-VIS~\cite{you2019hyperspectral} benchmark autonomous driving datasets demonstrate that our method achieves state-of-the-art performance, surpassing both RGB-based and hyperspectral segmentation approaches. In addition to quantitative gains, qualitative results show that our method generalizes well to diverse and complex driving environments, handling cluttered scenes, dense urban layouts, and visually similar material classes such as painted and unpainted metal.\looseness=-1

Our primary contributions are summarized as follows:
\begin{enumerate}[topsep=0pt,itemsep=0pt]
\item A novel hyperspectral adapter architecture tailored for semantic segmentation using HSI inputs.
\item A spectral transformer and spectrum-aware spatial prior module to jointly model spatial and spectral context.
\item A modality-aware interaction block that enables effective feature fusion between hyperspectral and pretrained foundational vision representations.
\item Comprehensive experiments and ablation studies across three benchmark datasets, demonstrating the effectiveness and generalization of our approach.
\item We release our code and pretrained models to facilitate further research in hyperspectral perception {\url{https://hsi-adapter.cs.uni-freiburg.de}}.
\end{enumerate}

\section{Related Work}
\label{sec:related-work}

We review existing work on hyperspectral semantic segmentation, including the challenges associated with extending these techniques to the hyperspectral domain.

{\parskip=2pt
\noindent\textit{Hyperspectral Semantic Segmentation}
is relatively underexplored in robotics compared to vision-based methods. Most existing work on hyperspectral segmentation originates from the remote sensing and geoscience fields, where datasets typically consist of a single aerial image covering urban or natural landscapes. These datasets offer limited variability and are often less relevant for robotics. More recently, novel hyperspectral datasets for urban autonomous driving have been introduced~\cite{gutierrez2023hsi, winkens2017hyko,you2019hyperspectral}.
Early approaches focus on processing the spectral dimension using dimensionality reduction techniques such as Principal Component Analysis (PCA) to tackle high-dimensional hyperspectral data~\cite{xue2021spectral}. However, these methods overlook spatial structure, which is crucial for accurate segmentation~\cite{li2019deep}. To overcome this limitation, subsequent work incorporates both spectral and spatial information through 3D convolutional neural networks~\cite{li2017spectral}. Although they improve performance, convolutional networks struggle to capture long-range dependencies in high-dimensional spectral data.

More recent work has introduced encoder-decoder architectures tailored for hyperspectral inputs, such as modified U-Net variants and DeepLab-based networks~\cite{gutierrez2023hsi,theisen2024hs3}. These methods achieve moderate improvements but they typically underperform compared to models trained on RGB representations, particularly when RGB inputs are synthesized from hyperspectral data. This performance gap is due to the use of network architectures and learning strategies tailored for RGB images, which do not fully leverage the spectral richness of hyperspectral inputs.
Transformer-based models have recently emerged as strong candidates for hyperspectral scene understanding. Their ability to model long-range dependencies makes them well-suited to the joint spatial-spectral structure of hyperspectral data. For instance, SpectralFormer~\cite{hong2021spectralformer} focuses on learning global relationships across spatial patches of hyperspectral images by embedding flattened patches and applying Transformer attention between them. In contrast, our proposed Spectral Transformer operates at the pixel level, modeling spectral dependencies directly across spectral bands without spatial patching. By preserving the original spatial structure, the Spectral Transformer reduces computational complexity, improves feature locality for semantic segmentation, and facilitates integration with multiscale feature extraction, further enhancing spatial feature quality.}

{\parskip=2pt
\noindent\textit{Adapters for Vision Foundation Models}: Transformer models, which form the backbone of modern vision foundation models, have achieved outstanding performance across multiple perception tasks. However, they often require significant computational resources for fine-tuning, especially for dense prediction~\cite{schramm2024bevcar,vodisch2024good}. Adapters offer a more efficient alternative to fine-tuning an entire transformer model by introducing small task-specific layers that are inserted into pretrained models~\cite{chen2022vision}. This enables efficient adaptation to new tasks without the need for full retraining, making it invaluable when annotations are scarce. The ViT-Adapter has been used to adapt vision transformers for dense prediction tasks such as semantic segmentation~\cite{hindel25balvit}, without requiring extensive retraining, thereby improving efficiency while maintaining performance.}

{\parskip=2pt
\noindent\textit{Adapters for Multimodal Segmentation}: Adapters have also been leveraged for knowledge transfer across modalities and tasks, particularly when training data is limited or modality-specific supervision is unavailable. By selectively fine-tuning small components that interact with large pretrained models, adapters enable efficient and flexible integration of additional input modalities without requiring full model retraining. MANet~\cite{ma2024manet} introduces a multimodal adapter to fine-tune the Segment Anything Model (SAM) by incorporating multi-scale geographic features from aerial imagery and LiDAR data. StitchFusion~\cite{li2024stitchfusion} presents a multi-directional adapter for combining RGB and infrared (IR) modalities, facilitating joint feature learning without the need for task-specific fusion layers. UniRGB-IR~\cite{yuan2024unirgb} proposes two specialized modules to enrich ViT representations with RGB and IR features. However, these methods typically operate on low-dimensional or single-channel modalities (e.g., grayscale IR), and their adapters are not designed to handle the high-dimensional spectral inputs found in HSI.}

\begin{figure*}
    \centering
    \includegraphics[width=0.79\linewidth]{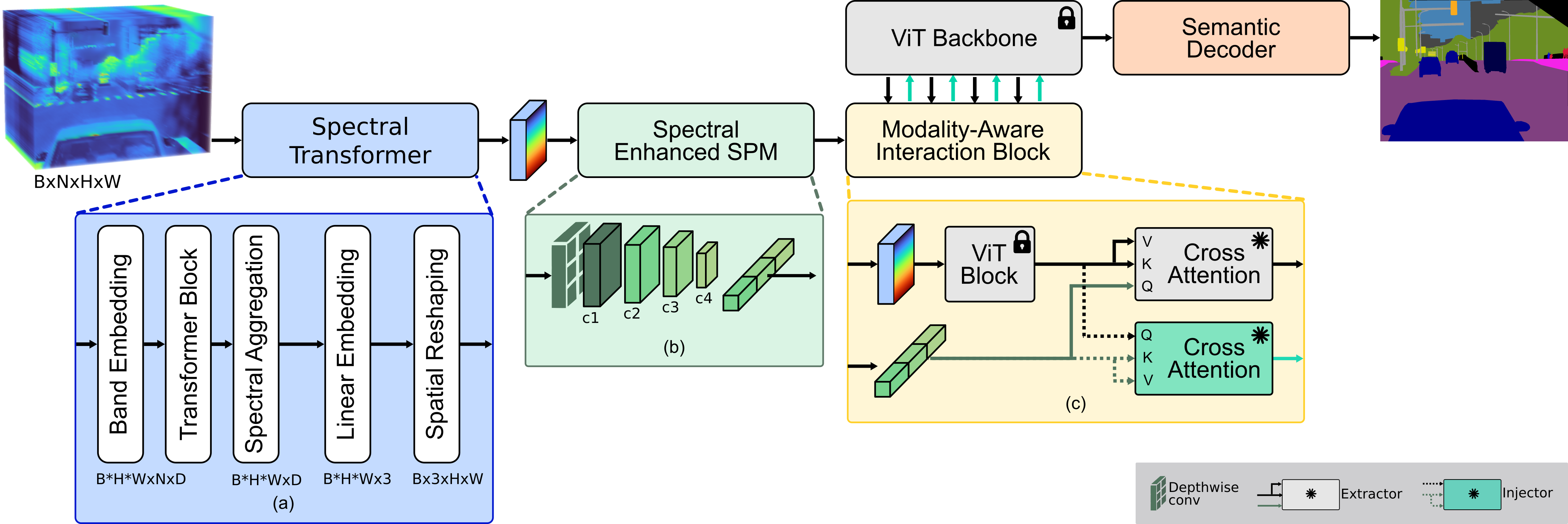}
    \caption{Our network takes hyperspectral images as input through the Spectral Transformer and a Spectral-Enhanced Spatial Prior Module (SPM), which extracts spectral and spatial features. These features interact with the frozen ViT backbone through a Modality-Aware Interaction Block, which utilizes gated bidirectional cross-attention. Finally, a semantic decoder yields pixel-level semantic class predictions from the fused representation.}
    \label{fig:arch}
   \vspace{-0.3cm}
\end{figure*}

More closely related to hyperspectral data, SAH-SCI~\cite{zeng2024sah} introduces a lightweight adapter framework for reconstructing hyperspectral images from compressed measurements, with a focus on spectral and spatial adaptation. However, their method targets reconstruction rather than semantic segmentation. In contrast, our hyperspectral adapter is specifically designed for dense segmentation. Unlike prior work based on grayscale infrared or compressed inputs, our method processes the complete hyperspectral cube and interacts with frozen RGB-pretrained vision transformers, enabling robust and data-efficient segmentation for robotic perception.

In summary, prior approaches to hyperspectral segmentation either adapt RGB-oriented architectures or design task-specific convolutional and transformer models, but fail to exploit the representational capacity of vision foundation models. Existing adapter-based frameworks, though effective in low-dimensional settings, are not designed for the high spectral dimensionality of HSI. We address this gap with a hyperspectral adapter that models spectral dependencies, incorporates spectrum-aware spatial priors, and enables interaction with frozen vision transformers, thereby leveraging foundation model generalization while accommodating the unique characteristics of hyperspectral data.
\section{Technical Approach}
\label{sec:technical-approach}

As shown in Fig.~\ref{fig:arch}, our proposed HSI-Adapter is a modular framework for hyperspectral semantic segmentation built upon vision foundation model backbones. It extracts spectral and spatial features through three network components: a Spectral Transformer, a Spectral-Enhanced Spatial Prior Module~(SPM), and a series of Modality-Aware Interaction Blocks. The Spectral Transformer captures long-range spectral dependencies and reduces channel dimensionality. The Spectral-Enhanced SPM integrates spatial and spectral information, while the interaction blocks enable bidirectional communication with the ViT through feature injection, fusion, and feedback.

We use a frozen ViT backbone initialized with weights obtained from unsupervised DINOv2 training~\cite{oquab2023dinov2} and supervised ImageNet training. 
Unlike standard ViTs, which operate with RGB images as inputs, our approach feeds hyperspectral features into the patch embedding module. The input is divided into non-overlapping $14\times14$ patches, flattened, projected into $D$-dimensional tokens, and passed through $L$ encoder layers. Simultaneously, in the HSI-Adapter the $N$-channel hyperspectral image is fed to the Spectral Transformer and refined by the Spectral-Enhanced SPM into multi-scale $D$-dimensional feature maps at $\frac{1}{8}$, $\frac{1}{16}$, and $\frac{1}{32}$ resolutions. These maps are flattened and concatenated, then input to the interaction modules. The ViT backbone is divided into $V$ stages, with an interaction block inserted after selected stages. At those stages, hyperspectral features are injected via deformable attention, fused with ViT tokens through gating, and enriched via bottleneck cross-attention. After $V$ interactions, the ViT tokens are reshaped into multi-scale feature maps, and a transposed convolution upsamples the feature map from $\frac{1}{8}$  to $\frac{1}{4}$ resolution for final segmentation predictions. Unless stated otherwise, the ViT patch embedding always receives a 3-channel image produced by the Spectral Transformer, and all ViT weights, including patch embedding and positional embeddings, remain frozen.

\subsection{Spectral Transformer}

The Spectral Transformer is designed to capture inter-band dependencies across the spectral dimension of HSI data. Unlike conventional Transformer architectures that employ attention over spatial tokens, this module operates on spectral vectors at each spatial location, treating the hyperspectral channels as a sequence. This formulation enables the model to learn complex, context-aware relationships between spectral bands, which are essential for accurate material and surface classification in HSI.

Given an input hyperspectral image $\mathbf{X} \in \mathbb{R}^{ N \times H \times W}$, where $N$ is the number of spectral bands and $(H, W)$ are the spatial dimensions, each spectral band at a given pixel is treated as an independent token. As shown in Fig.~\ref{fig:arch}a, each band is first projected into a $d_s$-dimensional feature space through a Band Embedding layer, producing a sequence of embedded tokens. This sequence is then processed by a Transformer encoder comprising $L_s$ layers and $h$ attention heads to model inter-band dependencies. Following spectral modeling, we apply average pooling across the spectral dimension to aggregate information, resulting in a representation $\mathbf{F} \in \mathbb{R}^{(HW) \times d_s}$. Finally, $\mathbf{F}$ is projected into a three-channel representation using a linear layer and then reshaped back into image space. This design enables the model to explicitly capture spectral dependencies at each spatial location while preserving the spatial resolution needed for dense prediction tasks.

\subsection{Spectral-Enhanced Spatial Prior Module}

Our Spectral-Enhanced Spatial Prior Module is tailored to extract multi-scale spatial features while leveraging spectral context captured during earlier processing stages. Compared to conventional spatial prior modules, our approach introduces an initial spectral-aware filtering step, followed by a hierarchical spatial encoding pipeline.
As illustrated in Fig.~\ref{fig:arch}~(b), the module employs a depthwise separable $3{\times}3$ convolution across spectral channels, followed by batch normalization and ReLU activation. This operation enhances local spectral contrast and reduces channel-wise noise prior to spatial feature extraction. Following spectral processing, we use a convolutional stem composed of three stacked $3{\times}3$ convolutions with batch normalization and ReLU activation, followed by a $3{\times}3$ max-pooling layer to downsample spatial resolution, yielding the first feature map $c_1$. To build a hierarchical feature pyramid, we employ three additional convolutional stages, each consisting of a $3{\times}3$ convolution with stride 2, batch normalization, and ReLU activation, following common design practices as in the spatial prior approach of~\cite{chen2022vision}. These stages generate feature maps $c_2$, $c_3$, and $c_4$, corresponding to downsampling factors of $8{\times}$, $16{\times}$, and $32{\times}$, respectively. We project the feature maps $c_2$, $c_3$, and $c_4$ into a shared embedding dimension $D$ using $1{\times}1$. Finally, these multiscale representations are fed to the adapter interaction blocks.\looseness=-1
 
\subsection{Modality-Aware Interaction Block}

Our Modality-Aware Interaction Block serves as a core component of the Hyperspectral Adapter, enabling bidirectional fusion between hyperspectral features and ViT tokens Fig.~\ref{fig:arch}~(c). Unlike prior approaches, our block introduces a dynamic, modality-aware fusion mechanism that adaptively integrates hyperspectral information while preserving the pretrained ViT structure. The interaction begins with an injector module that injects hyperspectral features extracted from the Spectral-Enhanced SPM into the ViT token stream using deformable cross-attention across multiple spatial levels. This process leverages learned reference points and multi-scale feature maps to perform efficient fusion.
We fuse the injected tokens with the original ViT tokens using a modality gating mechanism, which computes a dynamic per-token weighting:
\begin{align}
x_{out} = \gamma \cdot x_{\text{injected}} + (1 - \gamma) \cdot x_{vit},
\end{align}
where $x_{vit}$ are tokens before injection, $x_{\text{injected}}$ are tokens after deformable injection, and $\gamma \in [0, 1]$ is a learned per-token coefficient. 

This mechanism enables the model to blend injected features with the original ViT tokens in a content-aware manner, preserving the pretrained token distribution while integrating modality-specific cues. We then concatenate the fused representation with the class token and pass it through frozen ViT blocks, allowing the pretrained backbone to refine the fused tokens without updating its weights. Subsequently, the extractor stage uses deformable cross-attention in the reverse direction—attending from the updated ViT tokens back to the hyperspectral adapter features, to enhance the semantic richness of these features. To complete the interaction, a cross-attention feedback block further refines the adapter features based on the updated ViT context, improving alignment and ensuring consistency across modalities. In all interaction blocks, we use 3 feature levels {1/8, 1/16, 1/32} with 8 heads and 4 sampling points per head per level. Together, these components enable effective bidirectional interaction between hyperspectral priors and ViT representations, delivering strong semantic performance while maintaining compatibility with frozen pretrained backbones.

\subsection{Decoder}
We employ a UPerHead decoder to aggregate multi-scale features from four stages of the HSI-Adapter using pyramid pooling~\cite{xiao2018unified}. We then feed the fused representation through a convolutional head to yield the final segmentation logits. Additionally, we attach an auxiliary FCN head to an intermediate feature map to provide deep supervision during training.
We define the training loss as a weighted sum of the segmentation decoder loss and the auxiliary loss:
\begin{align}
\mathcal{L}_{\text{total}} = \mathcal{L}_{\text{seg}} + 0.4 \mathcal{L}_{\text{aux}},
\end{align}
where both $\mathcal{L}_{\text{main}}$ and $\mathcal{L}_{\text{aux}}$ represent standard pixel-wise cross-entropy losses applied to the heads.

\section{Experimental Results}
\label{sec:experiments}
This section investigates how hyperspectral information can be effectively leveraged with pretrained vision foundation models. Specifically, we ask: (1) Can pretrained RGB architectures be adapted to process hyperspectral inputs without training from scratch? (2) How can architectures be designed to enable compatibility between hyperspectral inputs and pretrained RGB encoders? (3) Does incorporating spectral-specific modules improve segmentation performance in challenging real-world scenes? We address these questions through experiments on three hyperspectral driving datasets, including detailed benchmarking, ablation studies, and qualitative analyses.

\begin{table*}[t]
    \footnotesize
    \centering
    \caption{Semantic segmentation results. The Input indicates whether the model was trained on full hyperspectral images (HSI) or projected RGB representations (pRGB) generated from HSI using~\cite{theisen2024hs3}. Reported metrics are mean Intersection-over-Union (mIoU), mean class accuracy (mAcc), and overall accuracy (aAcc), which are equivalent to the Acc\textsubscript{µ}, Acc\textsubscript{M}, F1\textsubscript{M}, and J\textsubscript{M} metrics used in~\cite{theisen2024hs3}. No test-time augmentation or multi-scale ensembling is applied. * denotes finetuning.}
    \renewcommand{\arraystretch}{1.1}
    \setlength{\tabcolsep}{4pt} 
    \begin{tabular}{>{\raggedright\arraybackslash}p{2.7cm}|
                    >{\centering\arraybackslash}p{1.2cm}|
                    c|
                    >{\centering\arraybackslash}p{0.70cm}
                    >{\centering\arraybackslash}p{0.70cm}
                    >{\centering\arraybackslash}p{0.70cm}|
                    >{\centering\arraybackslash}p{0.70cm}
                    >{\centering\arraybackslash}p{0.70cm}
                    >{\centering\arraybackslash}p{0.70cm}|
                    >{\centering\arraybackslash}p{0.70cm}
                    >{\centering\arraybackslash}p{0.70cm}
                    >{\centering\arraybackslash}p{0.70cm}|
                    >{\centering\arraybackslash}p{1.0cm}}
        \hline
        \multirow{2}{*}{\textbf{Method}} &
        \multirow{2}{*}{\textbf{Weights}} &
        \multirow{2}{*}{\textbf{Input}} &
        \multicolumn{3}{c|}{\textbf{HSI-Drive}} &
        \multicolumn{3}{c|}{\textbf{HCV2}} &
        \multicolumn{3}{c|}{\textbf{HyKo2}} &
        \textbf{\#Train} \\
        & & &
        \textbf{mIoU} & \textbf{aAcc} & \textbf{mAcc}
        & \textbf{mIoU} & \textbf{aAcc} & \textbf{mAcc}
        & \textbf{mIoU} & \textbf{aAcc} & \textbf{mAcc}
        & \textbf{Param} \\
        \hline
        U-Net  \cite{ronneberger2015u}           & x & HSI   & 64.45 & 94.95 & 74.74 & 37.73 & 85.25 & 48.62 & 57.39 & 85.36 & 68.15 & 31.2 \\
        RU-Net \cite{theisen2024hs3}            & x & HSI   & 67.11 & 96.08 & 76.92 & 42.23 & 87.63 & 54.14 & 58.64 & 86.72 & 68.79 & 34.5 \\
        RU-Net PCA \cite{theisen2024hs3}        & x & HSI   & 79.23 & 97.02 & 86.80 & 44.26 & 88.25 & 58.07 & 58.67 & 85.61 & 68.09 & 34.5 \\
        RU-Net  \cite{theisen2024hs3}           & x & pRGB  & 75.31 & 96.32 & 82.70 & 44.03 & 87.95 & 56.65 & 64.67 & 89.18 & 73.92 & 34.5 \\
        DL3+  \cite{chen2018encoder}            & IN-1K & HSI   & 56.63 & 67.86 & 65.58 & 40.79 & 86.60 & 53.15 & 53.22 & 84.10 & 63.01 & 45.7 \\
        DL3+ PCA \cite{theisen2024hs3}          & IN-1K & HSI   & 52.62 & 90.88 & 62.93 & 41.58 & 86.64 & 54.46 & 50.40 & 79.99 & 61.59 & 45.7 \\
        DL3+  \cite{chen2018encoder}            & IN-1K & pRGB  & 57.84 & 92.74 & 66.59 & 42.58 & 87.00 & 55.33 & 54.82 & 84.64 & 65.30 & 45.7 \\
        DL3+  \cite{theisen2024hs3}        & Cityscapes & pRGB  & 77.44 & 97.09 & 83.93 & 50.04 & 90.26 & 64.10 & 66.77 & 90.49 & 74.87 & 45.7 \\
        DL3+ * \cite{theisen2024hs3}        & Cityscapes & pRGB  & 73.84 & 95.69 & 81.95 & 48.47 & 89.62 & 61.91 & 65.41 & 88.62 & 73.97 & 45.7 \\
        ViT-Adapter \cite{chen2022vision}      & IN-1K & pRGB  & 65.42 & 95.00 & 72.90 & 43.29 & 86.56 & 55.81 & 53.61 & 79.60 & 65.11 & 52.6 \\
        ViT-Adapter \cite{chen2022vision}     & IN-22K & pRGB  & 65.82 & 94.89 & 73.67 & 43.78 & 86.71 & 56.52 & 56.42 & 89.60 & 72.95 & 52.6 \\
        \midrule
        HSI-Adapter (ours)    & IN-1k & HSI   & {93.63} & {99.24} & {96.16} & {56.25} & \underline{90.94} & {70.40} & {76.80} & \textbf{93.16} & {84.37} & 76.2 \\
        HSI-Adapter (ours)    & IN-22K & HSI   & \underline{93.75} &\textbf{99.27} & \underline{96.42} & \underline{56.38} & {90.55} & \underline{71.01} & \underline{76.90} & {93.05} & \underline{84.70} & 76.2 \\
        HSI-Adapter (ours)    & DINOv2 & HSI   & \textbf{93.80} & \textbf{99.27} & \textbf{96.47} & \textbf{58.81} & \textbf{91.54} & \textbf{72.48} & \textbf{77.14} & \underline{93.15} & \textbf{84.79} & 76.2 \\
        \hline
    \end{tabular}
    \label{tab:main_results}
\end{table*}

 \subsection{Datasets}
We evaluate our approach on three hyperspectral datasets that cover urban and rural driving scenarios: HSI-DriveV2~\cite{gutierrez2023hsi}, HyperspectralCityV2.0~\cite{you2019hyperspectral}, and HyKo2-VIS~\cite{winkens2017hyko}. In the following, we provide a detailed description of the datasets and their class distributions.

\textit{HSI-Drive v2.0}~\cite{gutierrez2023hsi} extends the original HSI-Drive~\cite{basterretxea2021hsi}, providing 752 manually annotated frames captured across all four seasons in real-world urban driving environments. HSI-DriveV2 focuses on semantic categories with challenging material compositions rather than standard semantic classes in autonomous driving. The dataset includes 10 semantic classes: road (60.73\%), road markings (3.02\%), vegetation (21.25\%), painted metal (2.16\%), sky (5.71\%), concrete (5.27\%), pedestrian (0.48\%), water (0.03\%), unpainted metal (0.79\%), and glass (0.56\%). Following the authors’ recommendation, we exclude the “water” class due to its extreme underrepresentation. The dataset features varying degrees of spectral variability across classes. Some classes exhibit similar spectral signatures, making them challenging to distinguish. For example, the “road” class consists solely of tarmac surfaces, leading to consistent spectral responses. In contrast, the “pedestrian” class includes diverse dynamic objects such as people, cyclists, motorcyclists, and animals, resulting in high intra-class variability. Data were acquired using a Photonfocus camera with a 25-band VIS-NIR filter spanning the visible to near-infrared spectrum, yielding hyperspectral cubes with a resolution of 1088$\times$2048$\times$25.

\textit{HyperspectralCityV2.0~(HCV2)}~\cite{you2019hyperspectral} comprises hyperspectral sequences captured in Shanghai over three consecutive days in June, under varying weather conditions (sunny and cloudy) and lighting scenarios including daytime, nighttime, and sunset. The dataset contains 367 training frames with coarse annotations and 55 testing frames with fine-grained semantic labels. The hyperspectral images span the 450–950~nm range, covering the visible to near-infrared spectrum. Each frame consists of 125 spectral channels, resulting in hyperspectral cubes of size 1400$\times$1800$\times$125. Semantic labels follow the Cityscapes class definition, including 19 categories such as road, sidewalk, pole, traffic light, traffic sign, and vegetation. However, eight of the original Cityscapes classes represent less than 1\% of the labeled data, and the “pole” class is entirely absent in the test set. Consequently, we exclude the “pole” class and focus on the remaining 18 categories in our experiments.

\textit{HyKo2-VIS}~\cite{winkens2017hyko} consists of hyperspectral images acquired in both urban traffic and rural road settings under driving conditions. The sensor operates entirely within the visible spectrum, covering the 470–630~nm range, and records 15 spectral bands per frame. Each frame is annotated with 10 semantic classes. The most prevalent categories are road (35.8\%), sky (15.2\%), grass (14.7\%), and vegetation (14.1\%), while less frequent classes include lane markings (1.1\%), panels (1.5\%), and person (0.03\%). Although HyKo2-VIS provides lower spectral and spatial resolution than HSI-DriveV2 and HCV2, it offers complementary diversity in geographic location, scene structure, and sensor characteristics.

\subsection{Experimental Setup}
We build our HSI-Adapter using a ViT-B backbone with deformable attention~\cite{zhu2020deformable}, using a patch size of 14. The adapter includes four feature interactions at transformer layer indexes [0, 2], [3, 5], [6, 8], and [9, 11]. In the final stage, we stack three multi-scale feature extractors to enhance semantic understanding across different resolutions. The vision transformer is initialized with a pretrain size of 518 and takes hyperspectral input with $N$ channels, depending on the dataset. The spatial transformer module within the adapter uses an embedding dimension of 32 and 4 attention heads for HS-Drive and HyKo2 and an embedding dimension of 160 and 5 attention heads for HCV2. For deformable attention, we use the configuration as described in~\cite{chen2022vision}. We train with a batch size of 2 and use a cosine learning rate scheduler with 1500 warm-up iterations. We use a learning rate of 0.0001 for HS-Drive and HCV2, and 0.00005 for HyKo2. We apply random cropping during training, with a crop size of (224, 448) for HS-Drive and HyKo2, and (448, 896) for HCV2. We do not use test-time augmentation or multi-scale evaluation.

\subsection{Benchmarking Results}

Tab.~\ref{tab:main_results} reports the performance in terms of mean Intersection-over-Union (mIoU), which is the primary evaluation metric, along with overall pixel accuracy (aAcc) and mean class accuracy (mAcc). We compare our method against both hyperspectral segmentation models and strong RGB-based baselines, including U-Net~\cite{ronneberger2015u}, DeepLabv3+~\cite{chen2018encoder}, and ViT-based RGB adapters~\cite{chen2022vision}. For HSI-based models, we include methods trained directly on hyperspectral inputs, with or without dimensionality reduction using PCA, with models such as RU-Net and DeepLabv3+.

\begin{table*}[t]
\centering
\footnotesize
\caption{Ablation on HSI-DriveV2. Variants (superscripts):
$^{a}$ ViT-B scratch HSI;
$^{b}$ ViT-B scratch pRGB;
$^{c}$ ViT-Adapter scratch HSI;
$^{d}$ ViT-Adapter RGB (scratch, ImageNet 1k, or 22k init);
$^{d}$ HSI-Adapter (ours): Spectral Transformer → 3-ch, Spectral-enhanced SPM at {1/8, 1/16, 1/32}, Modality-Aware Interaction Blocks (8 heads, 4 sampling points per head per level).
\textbf{Notes:} ViT-B/14 frozen incl. patch embed and positional embeddings; token dim $D\!=\!768$; single-scale, no test-time aug. HCV2 uses 18 classes since “pole” is absent; all baselines follow the same protocol.}

\label{tab:combined}
\begin{tabular}{@{}l p{2.0cm} c ccc p{1.0cm}@{}}
\toprule
\textbf{Method} & \textbf{Input} & \textbf{Pre-train} & \textbf{mIoU (\%)} & \textbf{mAcc (\%)} & \textbf{aAcc (\%)} & \textbf{Train~P (M)} \\
\midrule
25ch ViT$^{a}$ Backbone          & HSI   & $\times$    & 71.88 & 96.03 & 71.88 & 147.49 \\
3ch ViT$^{b}$ Backbone           & pRGB  & $\times$    & 62.52 & 93.10 & 73.34 & 144.18 \\
\midrule
25ch ViT-Adapter$^{c}$           & HSI   & $\times$    & 73.63 & 96.46 & 80.33 & 142.50 \\
3ch ViT-Adapter$^{d}$            & pRGB  & $\times$    & 65.21 & 94.50 & 73.43 & 139.20 \\
3ch ViT-Adapter$^{d}$            & pRGB  & IN-1k       & 65.42 & 95.00 & 72.90 & 52.623 \\
3ch ViT-Adapter$^{d}$            & pRGB  & IN-22k      & 65.82 & 94.89 & 73.67 & 52.623 \\
\midrule
Spectral Attention Prior   & HSI   & $\times$    & 69.57 & 95.65 & 77.61 & 52.614 \\
Spectral Attention Prior   & HSI   & IN-22k      & 87.54 & 91.94 & 98.10 & 52.614 \\
\midrule
Spectral Transformer       & HSI   & IN-22k      & 92.75 & 95.67 & 99.00 & 52.622 \\
+ Spectral-Enhanced SPM    & HSI   & IN-22k      & 93.61 & 96.09 & 99.24 & 52.623 \\
+ HSS-Interaction Blocks   & HSI   & IN-22k      & 93.75 & 99.27 & 96.42 & 76.246 \\
+ HSS-Interaction Blocks   & HSI   & DINOv2      & 93.80 & 99.27 & 96.47 & 76.246 \\
\bottomrule
\end{tabular}
\end{table*}

\begin{table*}[t]
\caption{Per-class IoU across adapters on HSI-DriveV2. We compare ViT-Adapter~\cite{chen2022vision} and our proposed network, both initialized with weights pretrained on ImageNet-22K.}
\centering
\setlength{\tabcolsep}{5.5pt}
\begin{tabular}{l|c|ccccccccc}
\toprule
Method & Pre-train & Road & Marks & Veg & P-Mtl & Sky & Concr & Ped & U-Mtl & Glass \\
\midrule
ViT-Adapter  & IN-22K   & 98.09 & 90.06 & 88.82 & 39.01 & 85.59 & 62.34 & 25.83 & 58.58 & 40.42 \\
HSI-Adapter (ours)  & IN-22K    & 99.78 & 98.73 & 98.51 & 88.54 & 97.99 & 92.87 & 97.9 & 83.07 & 86.37 \\
$\Delta$IoU & $\times$  & 1.58 & 7.87 & 9.99 & 43.18 & 10.22 & 31.72 & 68.16 & 23.33 & 38.83 \\
\bottomrule
\end{tabular}
\label{tab:per-class}
\end{table*}

We observe that our proposed HSI-Adapter significantly outperforms all competing methods in the mIoU score across the three benchmarks with all pretrained weights. Using DINOv2, our model in HSI-DriveV2, it achieves an mIoU of 93.80\%, exceeding the best-performing HSI baseline by 14.57\%. On the HCV2 benchmark, it achieves an mIoU score of 58.81\%, outperforming the closest HSI and RGB baselines by 14.55 and 8.77 points, respectively. Finally, on the HyKo2 dataset, it achieves an mIoU of 77.14\%, while also maintaining strong aAcc and mAcc values of 93.15\% and 84.79\%, respectively. To the best of our knowledge, this is the first time a hyperspectral segmentation model outperforms its RGB-based counterpart on multiple large-scale scene understanding benchmarks.

\subsection{Ablation Study}
\label{sec:ablation}

In this section, we analyze the progression of our HSI-Adapter architecture and evaluate different strategies for leveraging vision foundation models for hyperspectral semantic segmentation. Our goal is to quantify the contribution of each component in constructing a high-performing hyperspectral model and to understand the path toward surpassing RGB-trained variants built on pretrained backbones. We begin by training two baseline ViT models from scratch: one with 25 input channels for hyperspectral data, and another with three input channels using pseudo-RGB (pRGB) projections. The 25-channel model achieves nearly 10\% higher mIoU than its 3-channel counterpart, confirming the advantage of directly modeling hyperspectral inputs. However, this configuration is incompatible with vision foundation models, which are pretrained on standard 3-channel RGB images. Subsequently, we train adapter-based models with both 25-channel and 3-channel inputs. We train the 25-channel adapter from scratch entirely. For the 3-channel variant, we consider both training from scratch and initialization with ImageNet-1K, ImageNet-22K, or DINOv2 pretrained weights. Without pretraining, the 25-channel adapter outperforms the 3-channel model. Model variants and training settings for Tab.~\ref{tab:combined} are defined in the caption for reproducibility.


We present results of a Spectral Attention Prior, which consists of two key components: first, the SE block, which computes channel-wise attention using global average pooling followed by a fully connected layer with sigmoid activation; and second, a ViT-Adapter. This baseline enables the use of pretrained weights and the RGB-based adapter, which increases performance. Finally, we present ablation results on the complete HSI-Adapter architecture by progressively adding key components. Starting with the spectral transformer, we incrementally include the spectral-enhanced Spatial Prior Module (SPM) and the hyperspectral-specific interaction blocks. This final configuration achieves the highest performance, demonstrating the benefit of jointly modeling spectral and spatial information while effectively integrating pretrained semantic priors from the vision transformer.

\subsubsection{Per-Class IoU Analysis} To assess the class-specific effects of hyperspectral input, we compare the RGB-based ViT-Adapter and the hyperspectral-input HSI-Adapter in Tab.~\ref{tab:per-class}. The HSI-Adapter variant demonstrates consistent improvements across most classes, particularly in vegetation (+5.50\%), painted metal (+13.21\%), and pedestrian (+29.58\%), confirming the benefits of directly modeling hyperspectral cues even when pretrained weights remain fixed. These results indicate that integrating spectral-aware modules with strong pretrained features enables robust material recognition.

\begin{figure}[t]
    \centering
        \includegraphics[width=\linewidth]{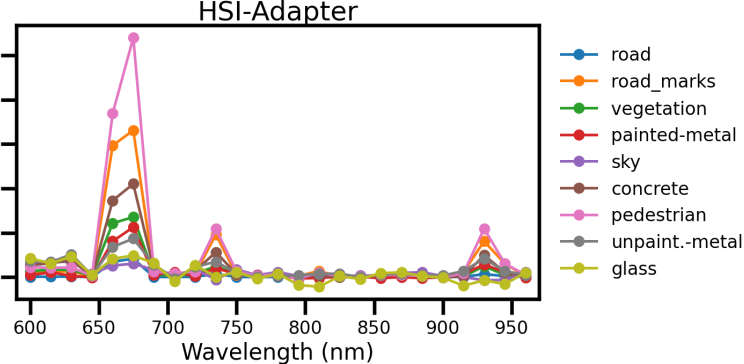}
    \caption{Per-band ablation results showing class-specific IoU drops ($\Delta$IoU) when zeroing individual spectral bands. The HSI-Adapter exhibits sharp spectral sensitivity, especially between 655–675\,nm. on the HSI-DriveV2 dataset.}
    \label{fig:bands_ablation}
    \vspace{-0.3cm}
\end{figure}

\subsubsection{Spectral Sensitivity Analysis}
To assess whether the model learns meaningful spectral representations, we perform a spectral ablation experiment where each of the 25 hyperspectral bands is individually zeroed out at test time. We report the corresponding IoU drop per class ($\Delta$IoU) in Fig.~\ref{fig:bands_ablation}. The HSI-Adapter demonstrates sharp, class-specific sensitivity to narrow spectral regions. Notably, zeroing bands in the 655--675\,nm range causes up to 25 percentage point drops for pedestrians and 10--15\,pp drops for painted metal and road markings. These results indicate that the Spectral Transformer exploits fine-grained reflectance variations that are lost in projection-based approaches, supporting its role in capturing discriminative spectral cues.

\subsection{Qualitative Results}
\begin{figure}[t]
    \centering
    \footnotesize
    \setlength{\tabcolsep}{1pt} 
    \begin{tabular}{cccc}
        Input & Improvement/Error & ViT-Adapter & HSI-Adapter (ours) \\
        
        \includegraphics[width=0.23\linewidth]{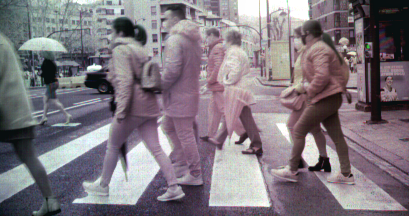} &
        \includegraphics[width=0.23\linewidth]{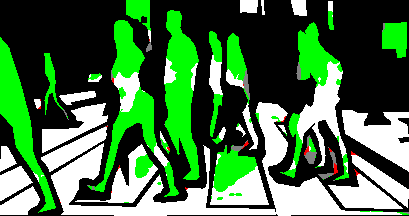} &
        \includegraphics[width=0.23\linewidth]{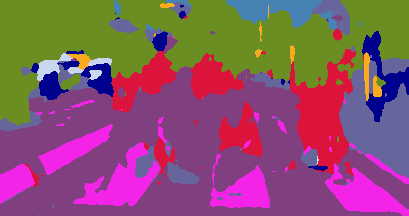} &
        \includegraphics[width=0.23\linewidth]{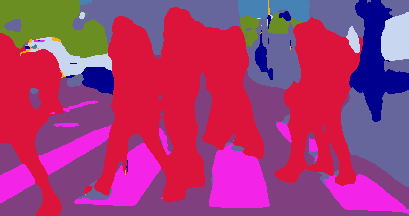} \\
        
        \includegraphics[width=0.23\linewidth]{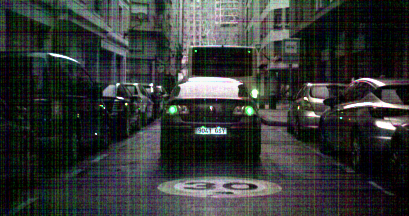} &
        \includegraphics[width=0.23\linewidth]{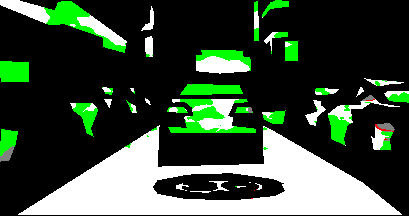} &
        \includegraphics[width=0.23\linewidth]{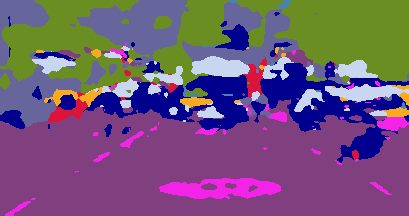} &
        \includegraphics[width=0.23\linewidth]{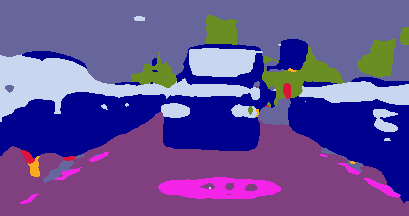} \\
        
    \end{tabular}
    \caption{Qualitative comparison on the HSI-DriveV2 dataset. Our proposed HSI-Adapter effectively segments challenging classes such as painted metal (dark blue), glass (light gray), and pedestrians (red) in scenes with significant visual ambiguity. In contrast, the baseline method fails to yield coherent segmentations, often predicting a mixture of incorrect classes across the scene. The Improvement/Error Map denotes pixels that are misclassified by our HSI-Adapter architecture in red, and the pixels that
ViT-Adapter architecture misclassifies but is correctly predicted by the HSI-Adapter model in green.}
    \label{fig:hsidrive_qualitative}
\end{figure}

\begin{figure}[t]
    \centering
    \footnotesize
    \setlength{\tabcolsep}{1pt} 
    \begin{tabular}{cccc}
        Input & Improvement/Error & ViT-Adapter & HSI-Adapter (ours) \\
        
        \includegraphics[width=0.23\linewidth]{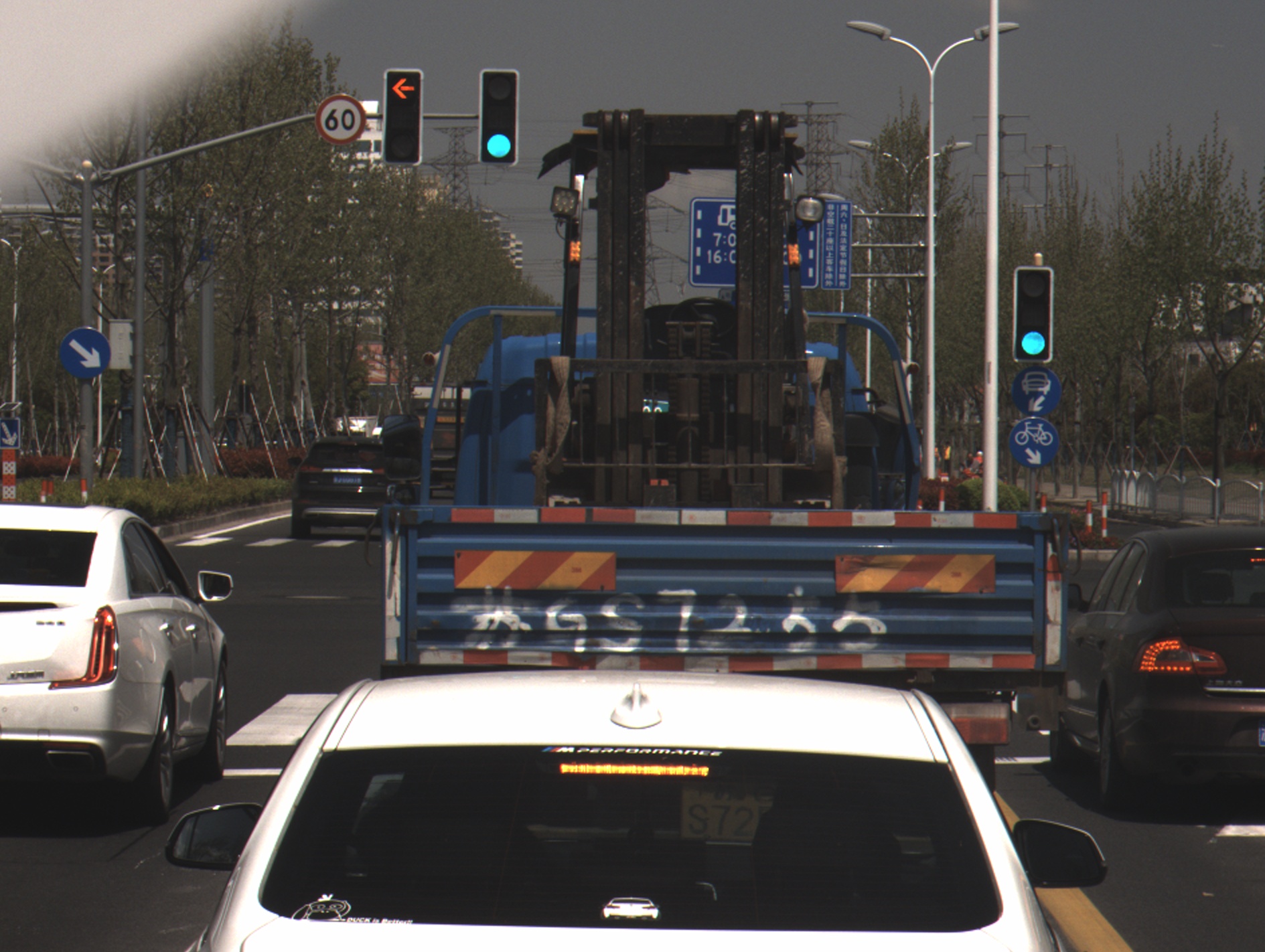} &
        \includegraphics[width=0.23\linewidth]{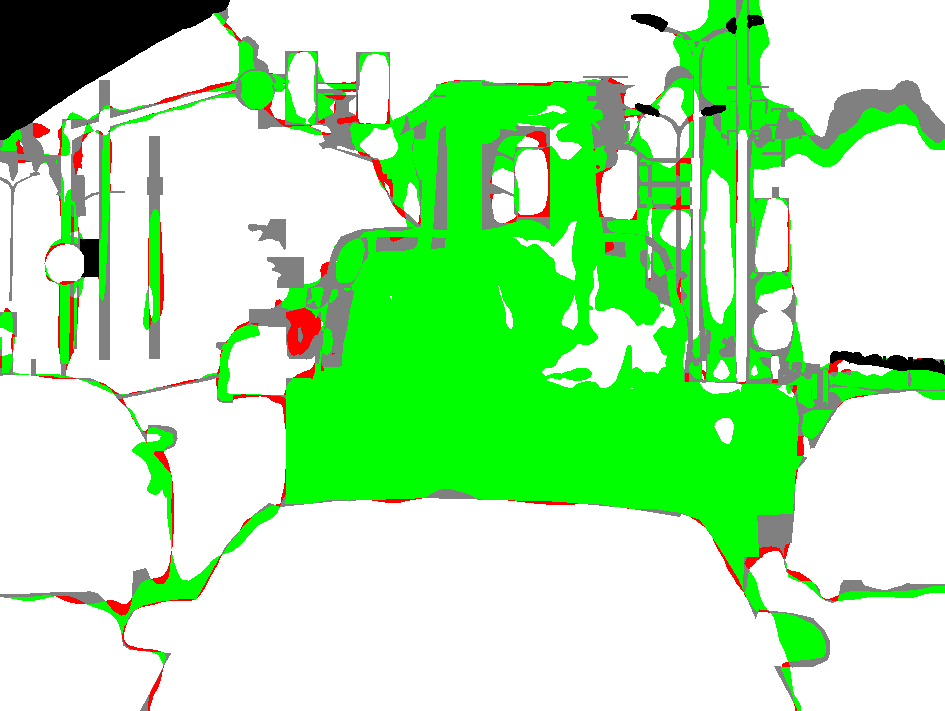} &
        \includegraphics[width=0.23\linewidth]{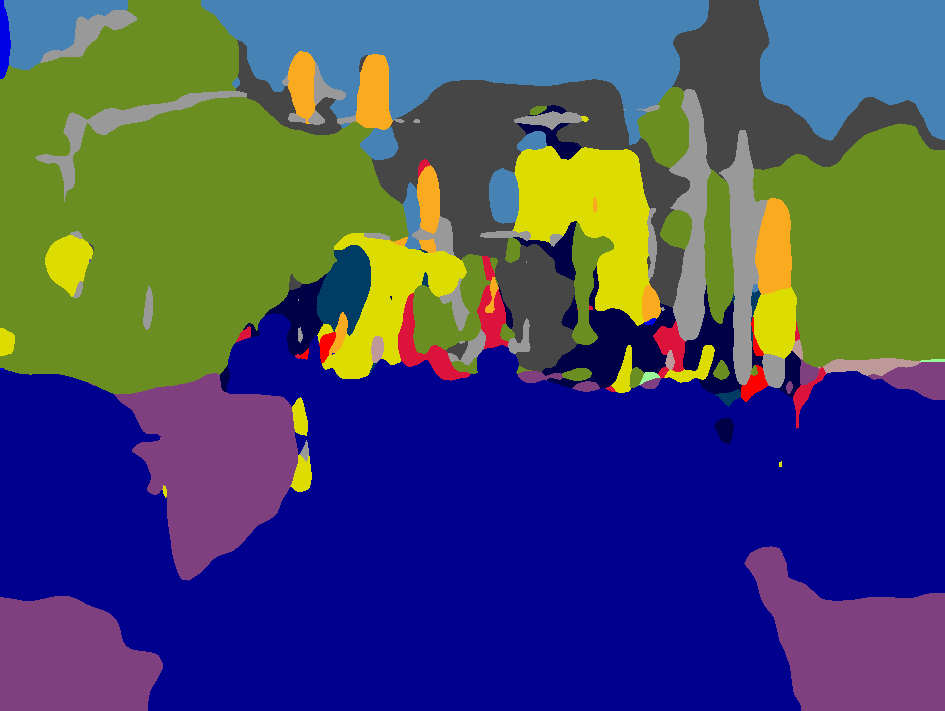} &
        \includegraphics[width=0.23\linewidth]{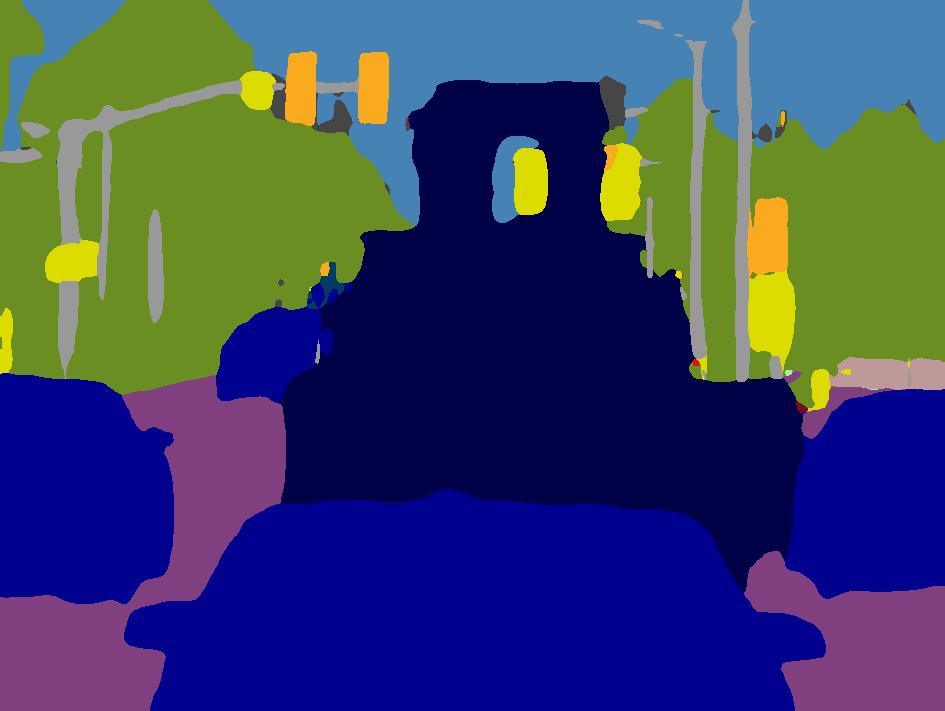} \\
        
        \includegraphics[width=0.23\linewidth]{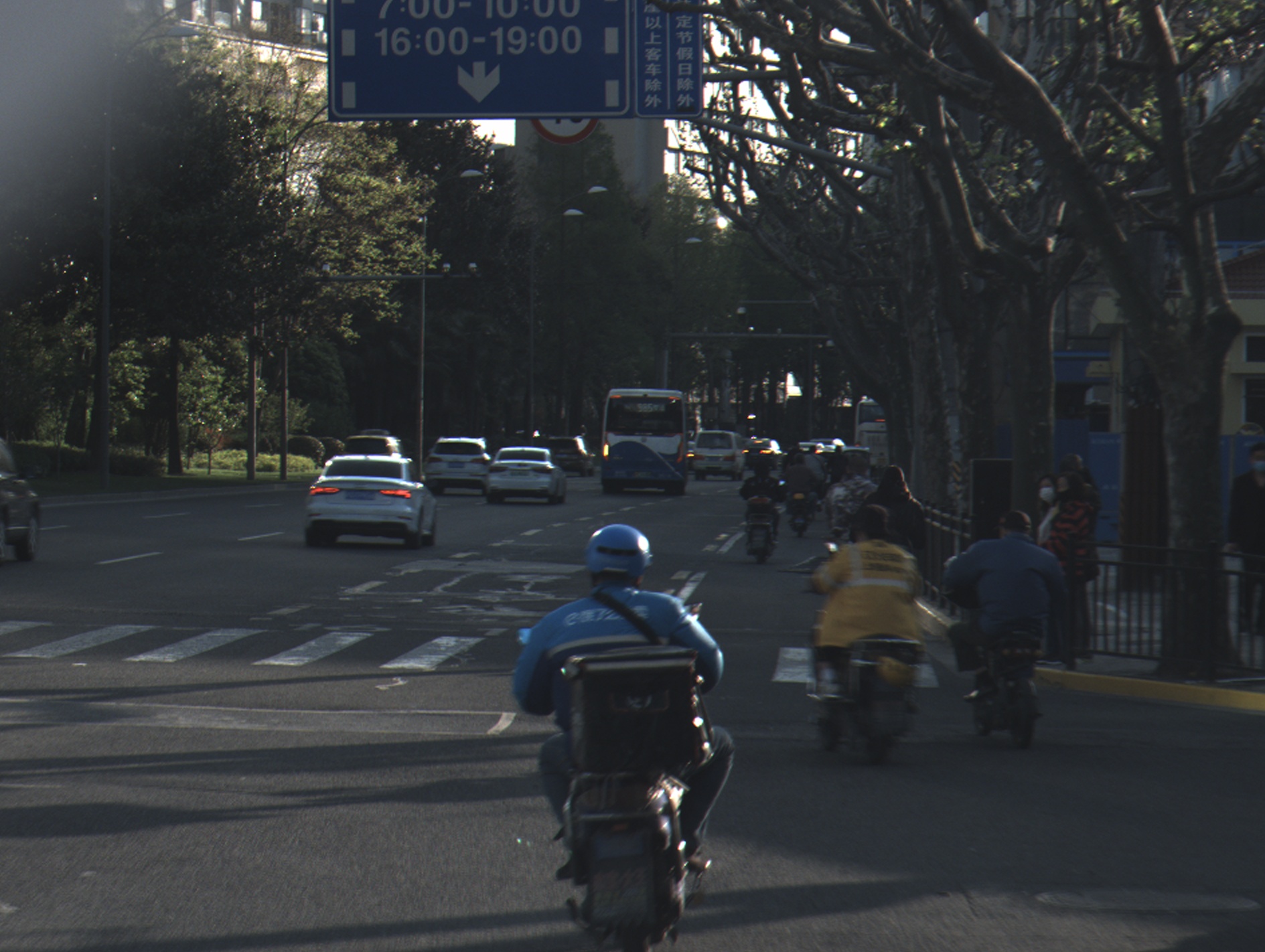} &
        \includegraphics[width=0.23\linewidth]{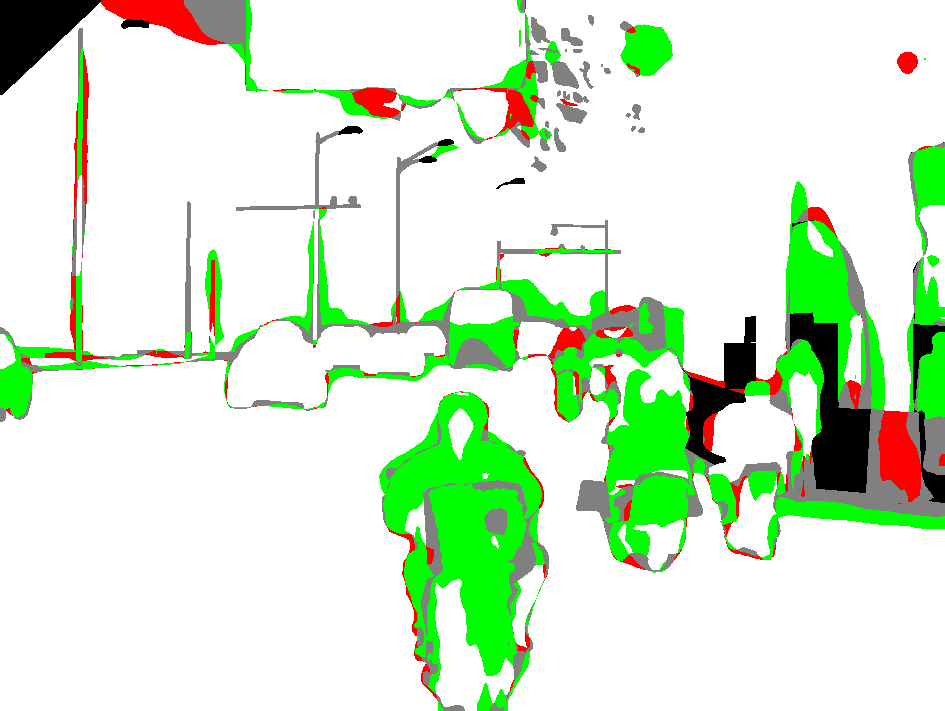} &
        \includegraphics[width=0.23\linewidth]{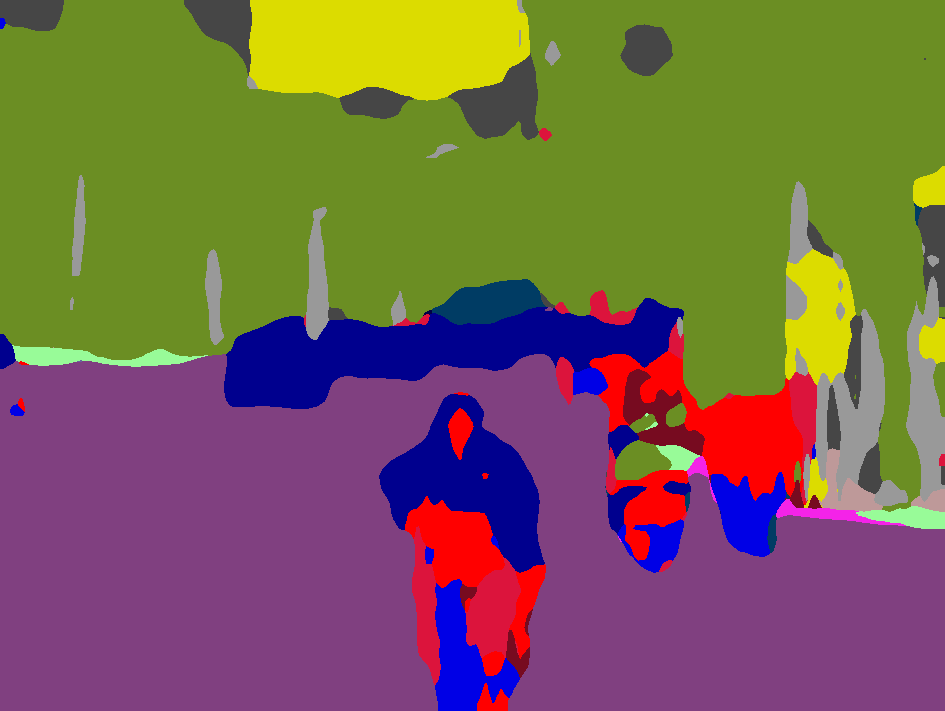} &
        \includegraphics[width=0.23\linewidth]{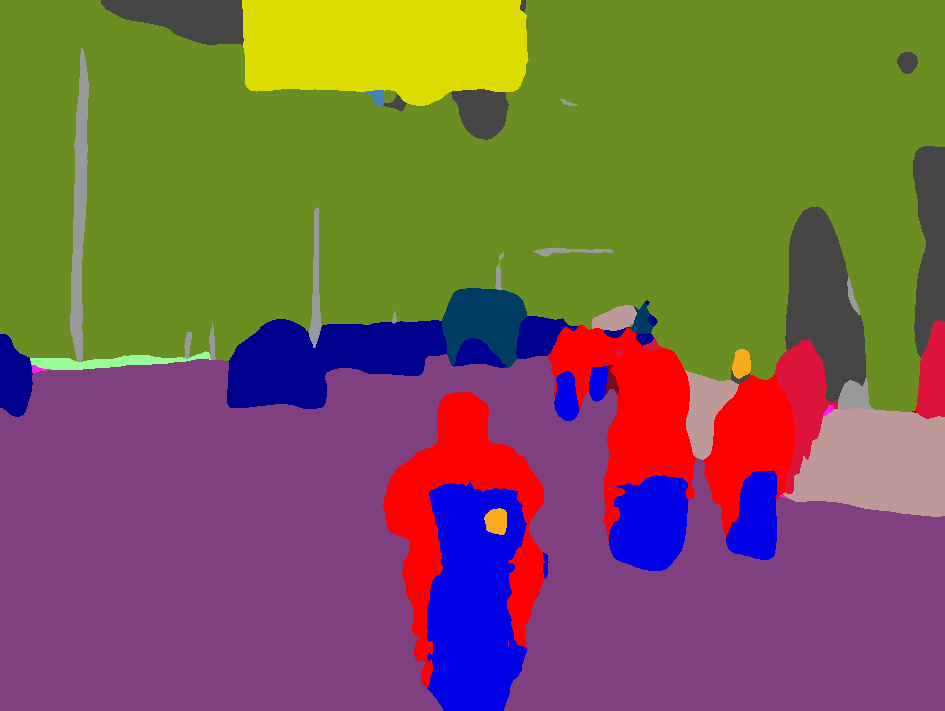} \\

    \end{tabular}
    \caption{Qualitative comparison on the HCV2 dataset. Our HSI-Adapter yields more accurate segmentation results, particularly in visually cluttered scenes. }
    \label{fig:hcv2_qualitative}
   \vspace{-0.5cm}
\end{figure}

Fig.~\ref{fig:hsidrive_qualitative}  presents qualitative results on the HSI-DriveV2 dataset. The first column shows the input hyperspectral images (visualized as pRGB), followed by an improvement/error map, predictions from the baseline ViT-Adapter model, and the outputs of our proposed HSI-Adapter. In the Improvement/Error column, green regions indicate areas where our HSI-Adapter outperforms the ViT-Adapter baseline, while red highlights its errors. Our method demonstrates consistent improvements in both complex urban scenes (top row) and highway environments (bottom row). Specifically, the HSI-Adapter more accurately delineates pedestrians and cars, capturing finer boundaries and reducing confusion between object classes such as painted and unpainted metal, roads, sidewalks, and vegetation. Compared to the ViT-Adapter, our HSI-Adapter yields cleaner segmentation maps with less noise and sharper object contours. This highlights the effectiveness of leveraging hyperspectral information through our proposed adapter architecture, particularly in scenarios with cluttered scenes (both rows) and challenging lighting conditions (bottom row).\looseness=-1

\begin{figure}[t]
    \centering
    \footnotesize
    \setlength{\tabcolsep}{1pt} 
    \begin{tabular}{cccc}
        Input & Improvement/Error & ViT-Adapter & HSI-Adapter (ours) \\
        
        \includegraphics[width=0.23\linewidth]{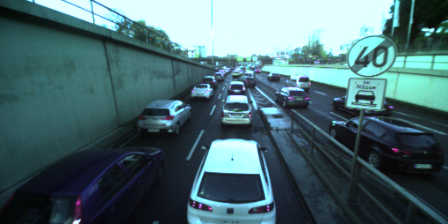} &
        \includegraphics[width=0.23\linewidth]{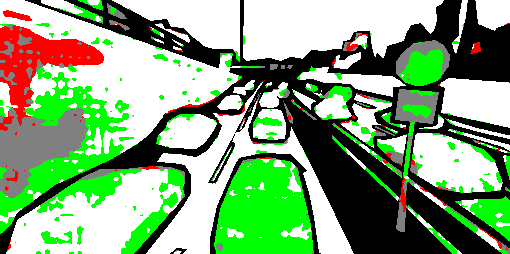} &
        \includegraphics[width=0.23\linewidth]{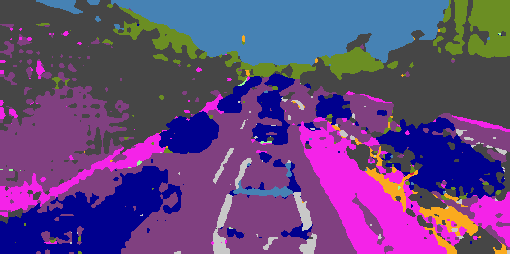} &
        \includegraphics[width=0.23\linewidth]{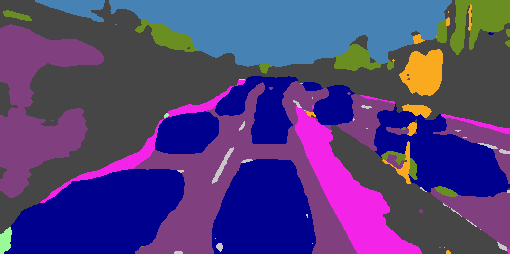} \\
        
        \includegraphics[width=0.23\linewidth]{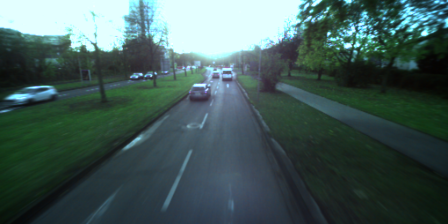} &
        \includegraphics[width=0.23\linewidth]{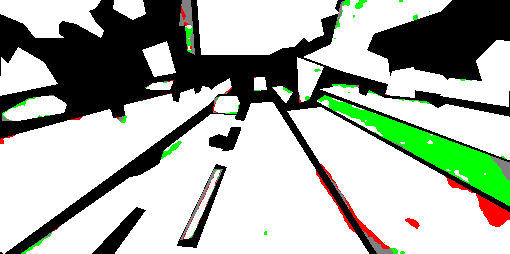} &
        \includegraphics[width=0.23\linewidth]{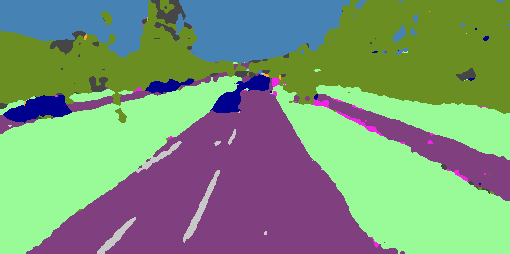} &
        \includegraphics[width=0.23\linewidth]{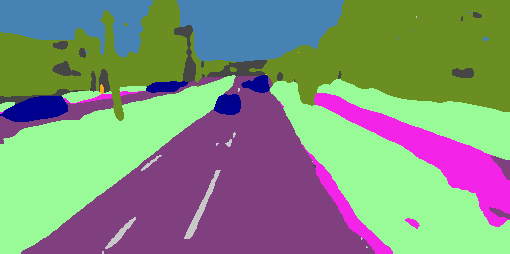} \\
        
    \end{tabular}
    \caption{Qualitative results on the HyKo2 dataset. Our method consistently achieves accurate segmentation results in dense urban environments with multiple vehicles (dark blue). Notably, the HSI-Adapter effectively distinguishes between visually similar classes such as road (purple) and sidewalk (magenta), whereas the baseline method often misclassifies them as the same category.}
    \label{fig:hyko_qualitative}
    \vspace{-0.3cm}
\end{figure}

Fig.~\ref{fig:hcv2_qualitative} shows qualitative results on the HCV2 dataset. In the top row, our HSI-Adapter accurately segments the truck even though it is partially occluded. The ViT-Adapter misclassifies the occluded truck into multiple incorrect classes. In the bottom row, our method correctly segments people (red), a motorcycle (blue), cars (dark blue), and a truck, despite the scene’s visual clutter and complex material composition. Additionally, our model provides more precise segmentation of poles (gray), traffic lights (yellow), and signs (orange), highlighting its effectiveness in capturing fine-grained structures.

In Fig.~\ref{fig:hyko_qualitative}, we present qualitative results on the HyKo2 dataset. In the top row, the HSI-Adapter provides significantly cleaner segmentation of vehicles and road markings, while the ViT-Adapter struggles with class confusion, particularly in the central lane and surrounding vehicles. In the bottom row, the HSI-Adapter yields more consistent lane boundary segmentation and cleaner separation of road, sidewalk, and vegetation areas. It also corrects misclassifications of buildings and sky that are present in the ViT-Adapter's output. These results show that the spectral information in HSI allows our adapter to resolve ambiguities that RGB-based models face, especially in cluttered scenes, fine structures, and visually similar classes.

\subsection{Failure Cases}

\begin{figure}[t]
    \centering
    \footnotesize
    \setlength{\tabcolsep}{1pt} 
    \begin{tabular}{cccc}
        Input & Improvement/Error & ViT-Adapter & HSI-Adapter (ours) \\
        \includegraphics[width=0.23\linewidth]{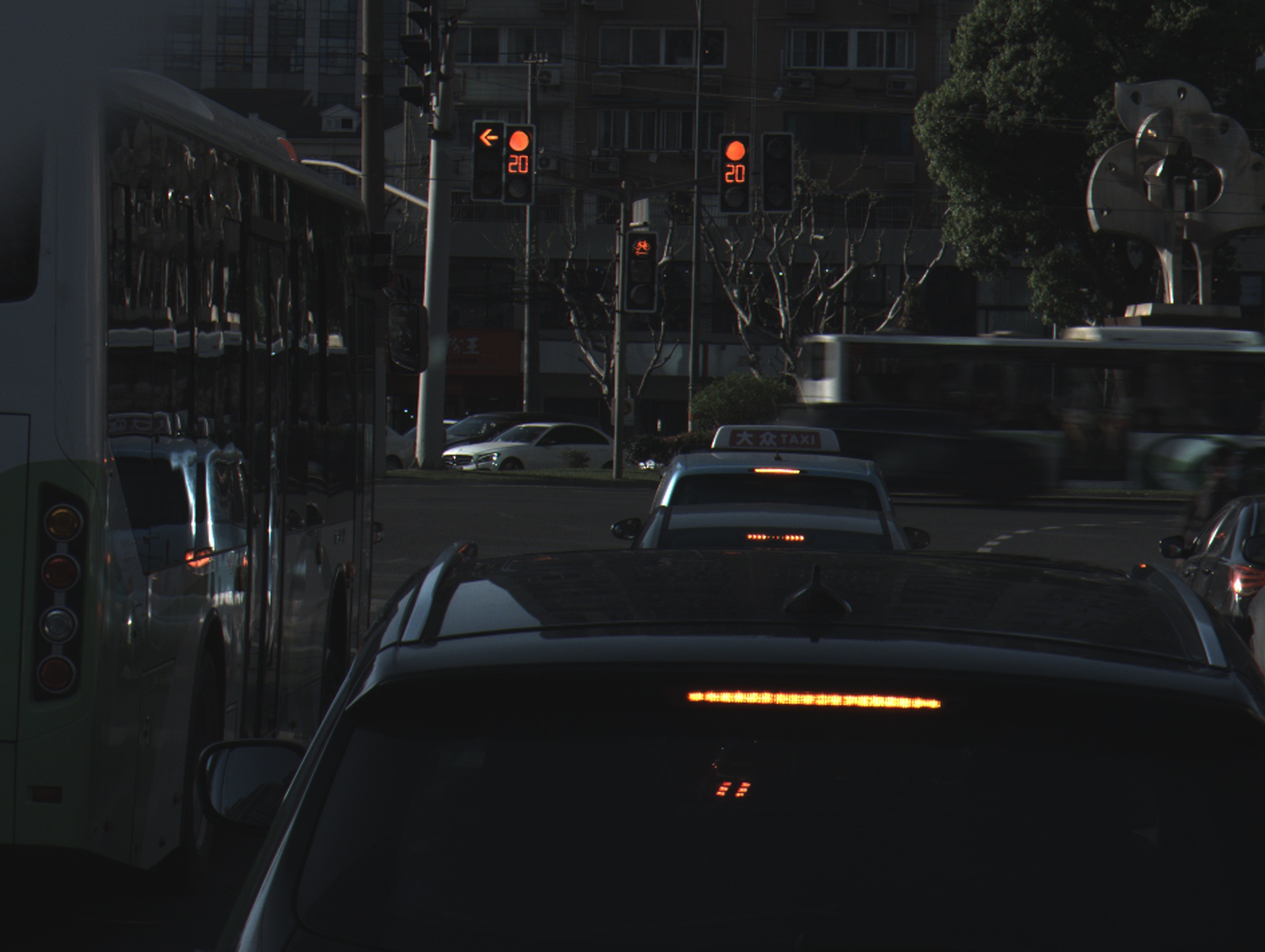} &
        \includegraphics[width=0.23\linewidth]{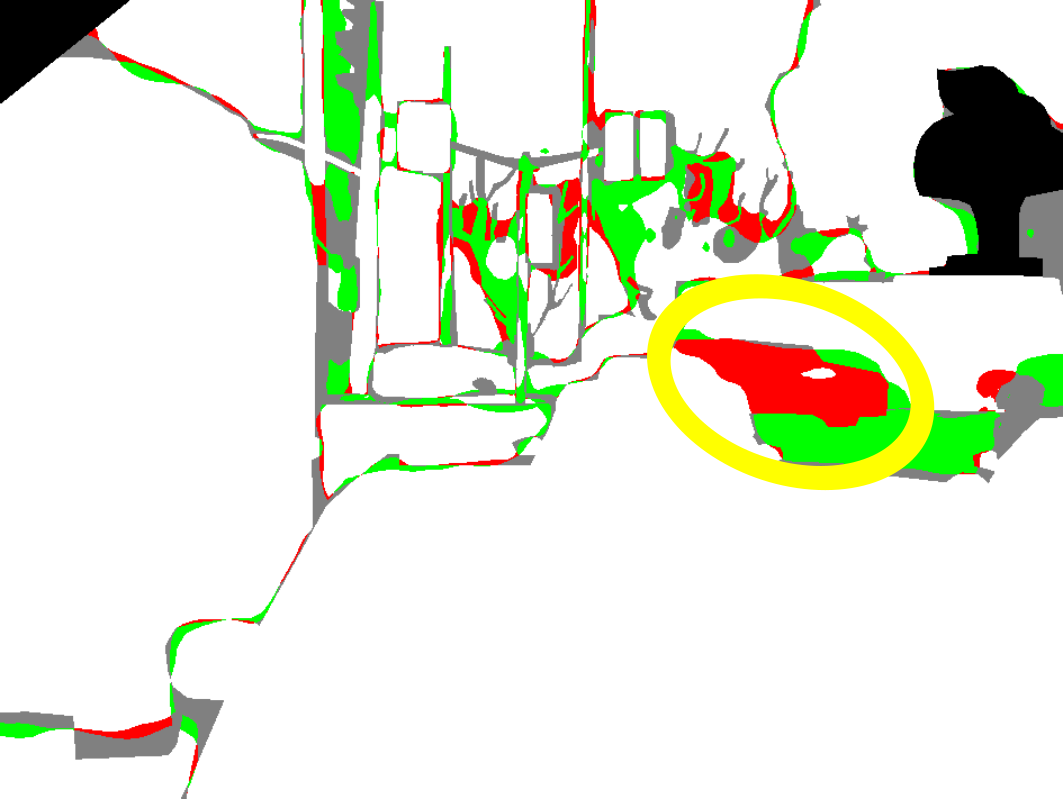} &
        \includegraphics[width=0.23\linewidth]{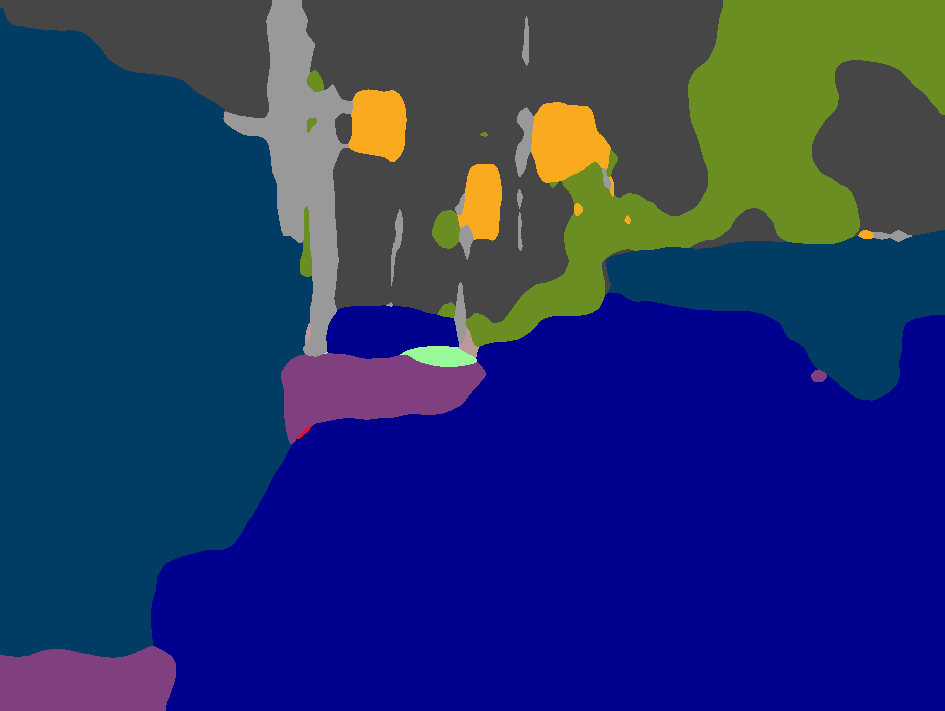} &
        \includegraphics[width=0.23\linewidth]{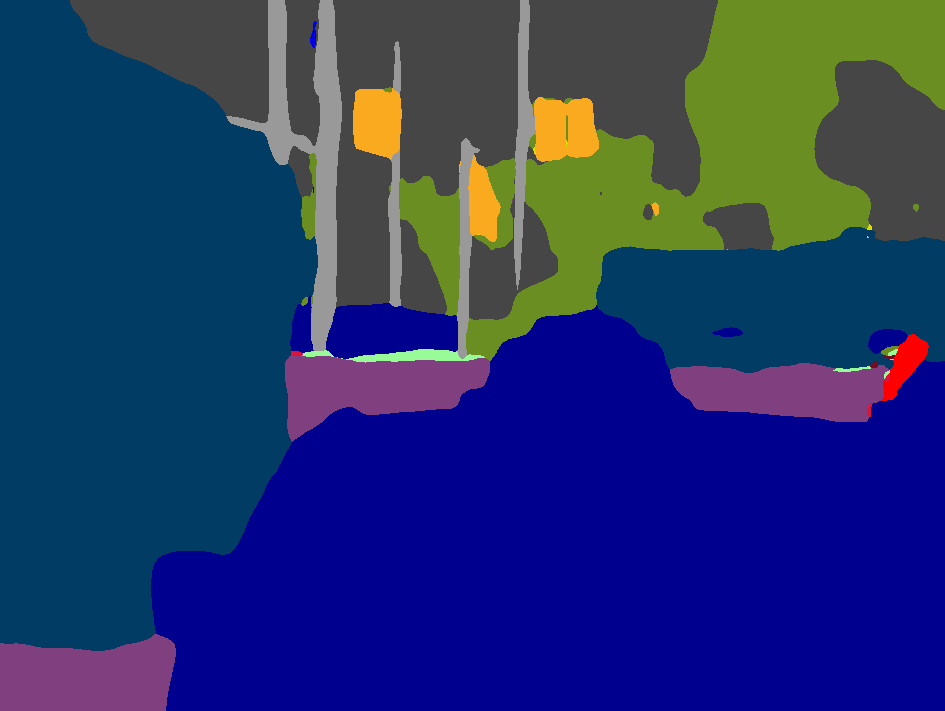} \\
        
        \includegraphics[width=0.23\linewidth]{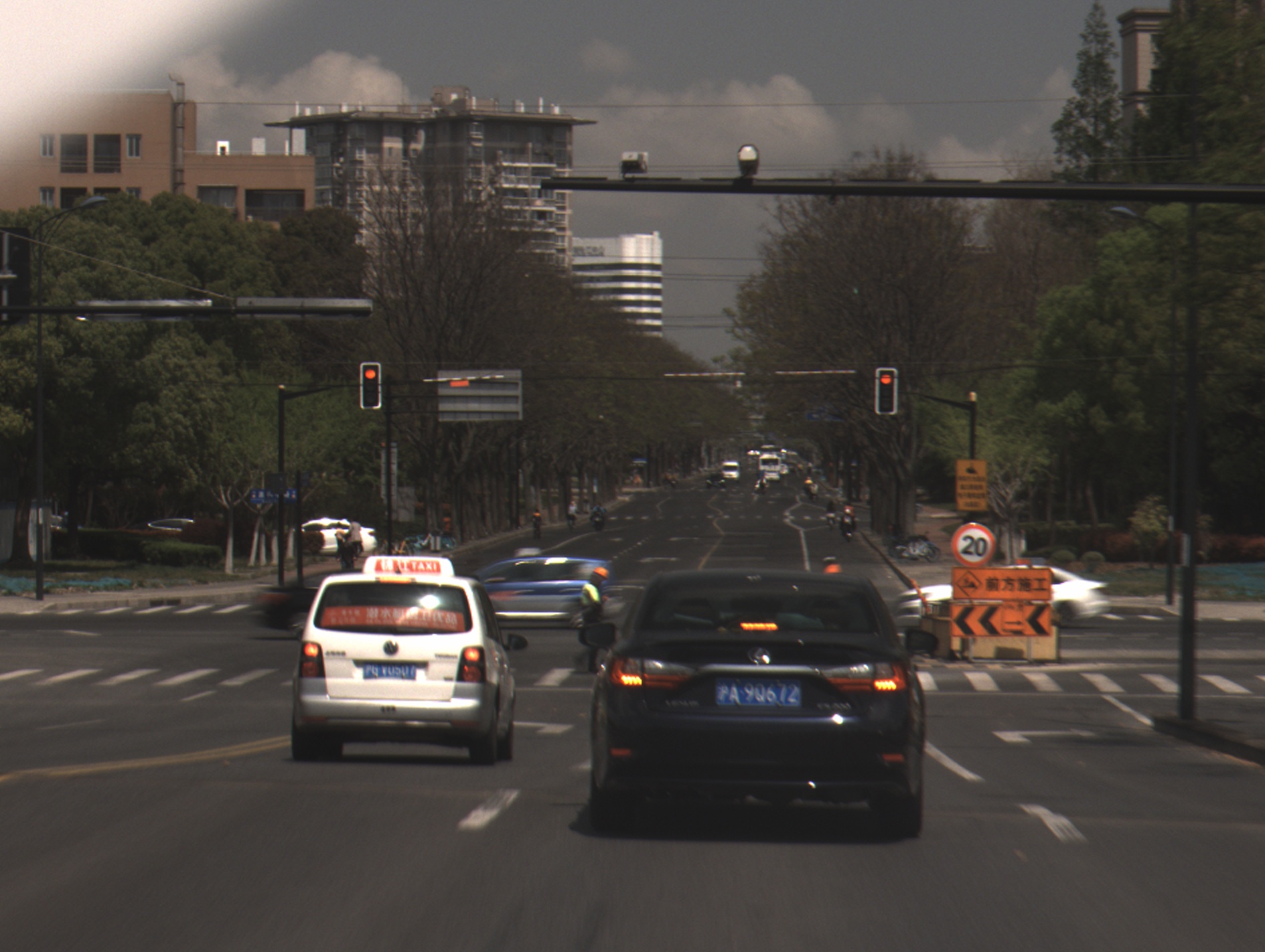} &
        \includegraphics[width=0.23\linewidth]{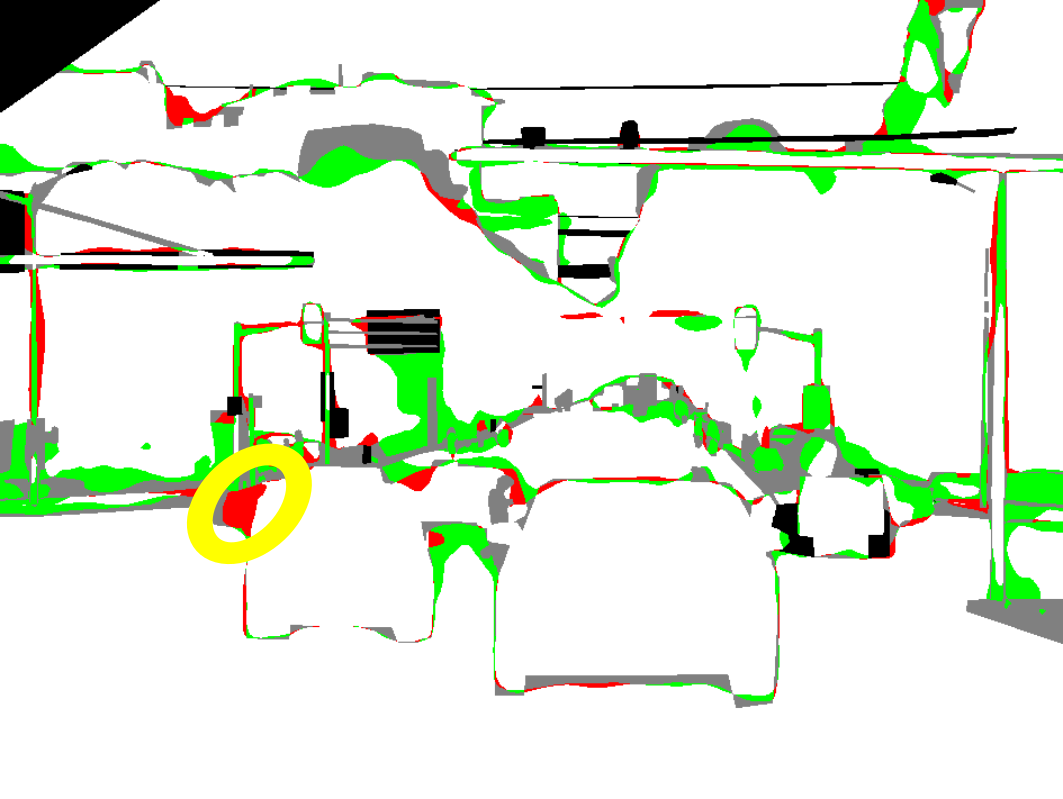} &
        \includegraphics[width=0.23\linewidth]{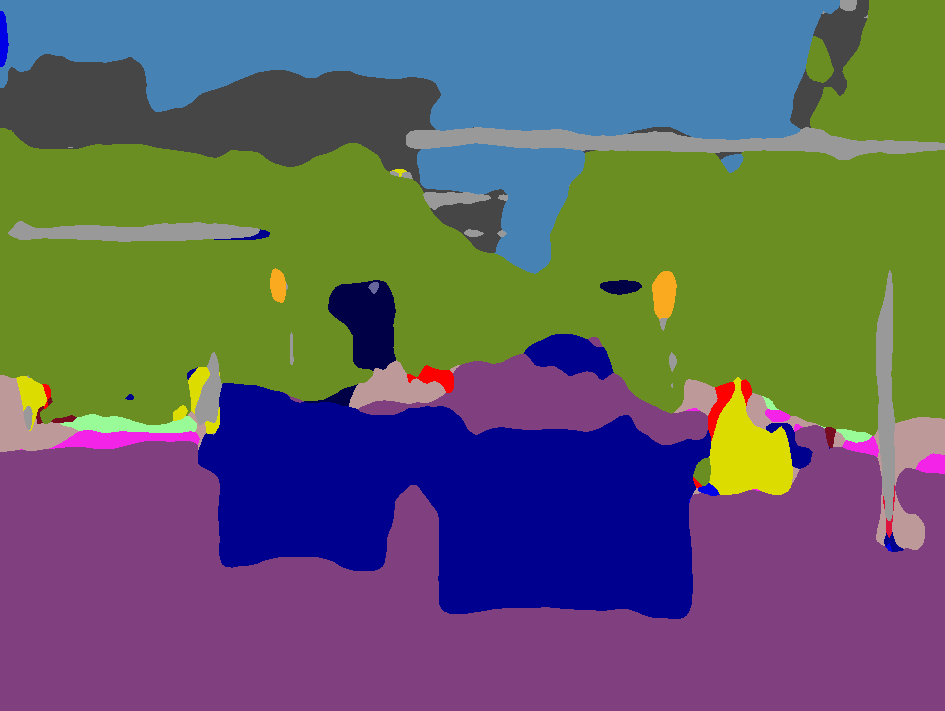} &
        \includegraphics[width=0.23\linewidth]{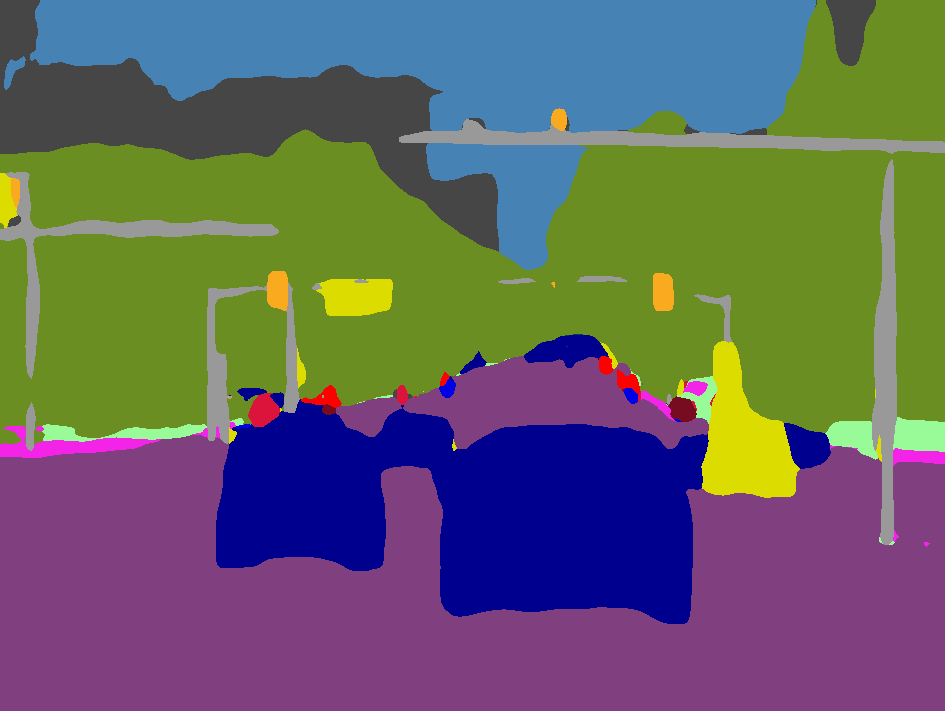} \\

        \includegraphics[width=0.23\linewidth]{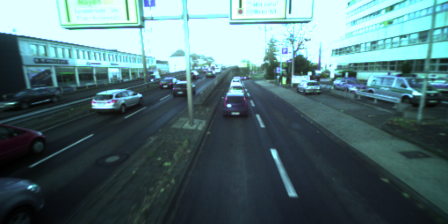} &
        \includegraphics[width=0.23\linewidth]{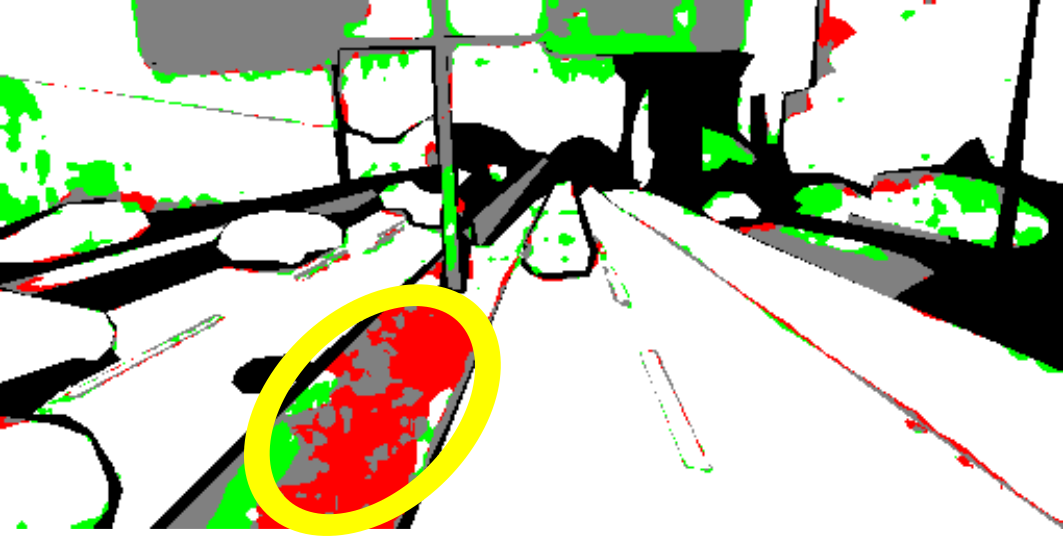} &
        \includegraphics[width=0.23\linewidth]{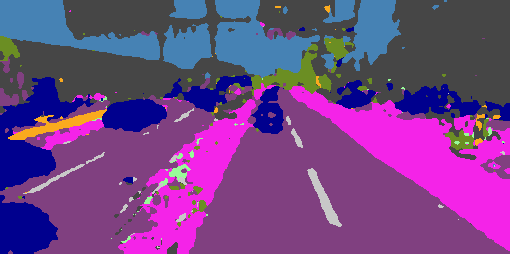} &
        \includegraphics[width=0.23\linewidth]{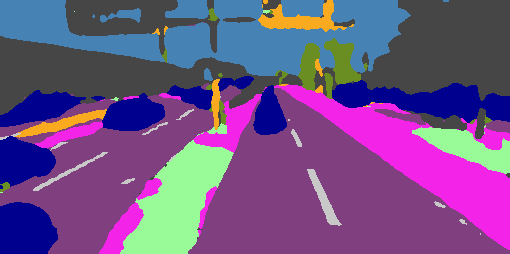} \\

    \end{tabular}
    \caption{Failure case examples with yellow circles highlighting regions where the HSI-Adapter underperforms compared to the ViT-Adapter. In the first two rows, our method fails to segment vehicles affected by motion distortion, resulting in missed segmentation. The third row presents a HyKo2 sample where the HSI-Adapter misclassifies portions of the sidewalk as vegetation due to overgrown weeds, illustrating a potential class mixture and labeling ambiguity.\looseness=-1}
    \label{fig:fail}
\end{figure}

We present failure cases in Fig.~\ref{fig:fail}. Our HSI-Adapter struggles particularly in scenarios heavily affected by motion blur in HCV2. In the first two rows, within the yellow-circled areas, the model fails to detect fast-moving cars which appear heavily distorted, while the ViT-Adapter is able to segment them more effectively. We hypothesize this is due to the inherent challenges of hyperspectral image acquisition. Hyperspectral sensors often require longer exposure times to capture detailed spectral information, making them especially susceptible to motion blur in dynamic scenes. Reducing the exposure time can help minimize blur but it results in underexposed and noisy images, particularly in low-light conditions. In contrast to RGB cameras, where the aperture can be adjusted to allow more light in, changing the aperture or the angle of light incidence in HSI systems affects the Fabry-Perot filters~\cite{gutierrez2023hsi}. This leads to variations in the spectral response and further hampers data acquisition. Since HCV2 provides RGB images with the hyperspectral data, the ViT-Adapter is not affected by these problems and is better able to preserve object structure in scenes with significant motion.

In the bottom row, we show a sample scene from HyKo2 where the HSI-Adapter misclassifies pixels labeled as sidewalk. Interestingly, the sidewalk in this scene has overgrown weeds, which our method interprets as vegetation. This example highlights a potential class mixture and labeling ambiguity that presents a challenge even for human annotators.
\section{Conclusion}
\label{sec:conclusion}

In this work, we proposed a novel hyperspectral adapter architecture that leverages vision foundation models for semantic segmentation with hyperspectral inputs. Our approach introduces a spectral transformer and a spectrum-aware spatial prior module that jointly model spatial and spectral dependencies. To bridge the gap between hyperspectral data and pretrained vision foundation model representations, we developed a modality-aware interaction block that enables rich cross-domain feature fusion. We performed extensive experiments and ablation studies across three challenging autonomous driving benchmarks and demonstrated that our HSI-Adapter generalizes well across diverse datasets, consistently outperforming both RGB-based and existing HSI-based methods. To the best of our knowledge, this is the first hyperspectral segmentation model to surpass its RGB counterpart while fully leveraging strong pretrained priors for autonomous driving perception.


\footnotesize
\bibliographystyle{IEEEtran}
\bibliography{references.bib}

\end{document}